# Reinforcement learning-based dynamic cleaning scheduling framework for solar energy system

Heungjo An[1]

Department of Industrial Engineering, Kumoh National Institute of Technology, Gumi City, South Korea

**Abstract**

Advancing autonomous green technologies in solar photovoltaic (PV) systems is key to improving sustainability and efficiency in renewable energy production. This study presents a reinforcement learning (RL)-based framework to autonomously optimize the cleaning schedules of PV panels in arid regions, where soiling from dust and other airborne particles significantly reduces energy output. By employing advanced RL algorithms, Proximal Policy Optimization (PPO) and Soft Actor-Critic (SAC), the framework dynamically adjusts cleaning intervals based on uncertain environmental conditions. The proposed approach was applied to a case study in Abu Dhabi, UAE, demonstrating that PPO outperformed SAC and traditional simulation optimization (Sim-Opt) methods, achieving up to 13% cost savings by dynamically responding to weather uncertainties. The results highlight the superiority of flexible, autonomous scheduling over fixed-interval methods, particularly in adapting to stochastic environmental dynamics. This aligns with the goals of autonomous green energy production by reducing operational costs and improving the efficiency of solar power generation systems. This work underscores the potential of RL-driven autonomous decision-making to optimize maintenance operations in renewable energy systems. In future research, it is important to enhance the generalization ability of the proposed RL model, while also considering additional factors and constraints to apply it to different regions.

## 1. Introduction

Solar PV systems rely heavily on solar irradiance for power generation, making arid regions with high irradiance levels prime locations for solar farms. However, these regions face a critical challenge: the accumulation of airborne particles such as dust and sand on the surface of PV panels. This phenomenon, known as soiling, includes not only dust but also dirt, organic debris, and bird droppings [1]. Soiling significantly reduces the efficiency and output of PV systems by obstructing sunlight.

Local weather plays a crucial role in the soiling process by influencing both the deposition and removal of particles on PV panels. Key environmental factors such as airborne dust, wind speed, humidity, temperature, and rainfall significantly impact PV panel efficiency. Among these, airborne dust is the primary contributor to soiling, while wind has a dual effect: it promotes dust accumulation at low speeds and facilitates dust removal at higher speeds. These factors collectively determine the extent of dust buildup on PV surfaces, which can obstruct sunlight, reduce irradiance, and consequently decrease the power output of the panels.

Consequently, regular cleaning of panel surfaces becomes a critical managerial operation to restore panel efficiency in arid regions. For example, Figure 1 illustrates the soiling and cleaning cycle of solar PV panels in such environments. The top row shows the process: a sandstorm deposits dust and particles on the panels (soiling), which reduces their efficiency, followed by a cleaning process that restores the panels to their optimal state.

---

[1] Correspondence: heungjo.an@kumoh.ac.kr

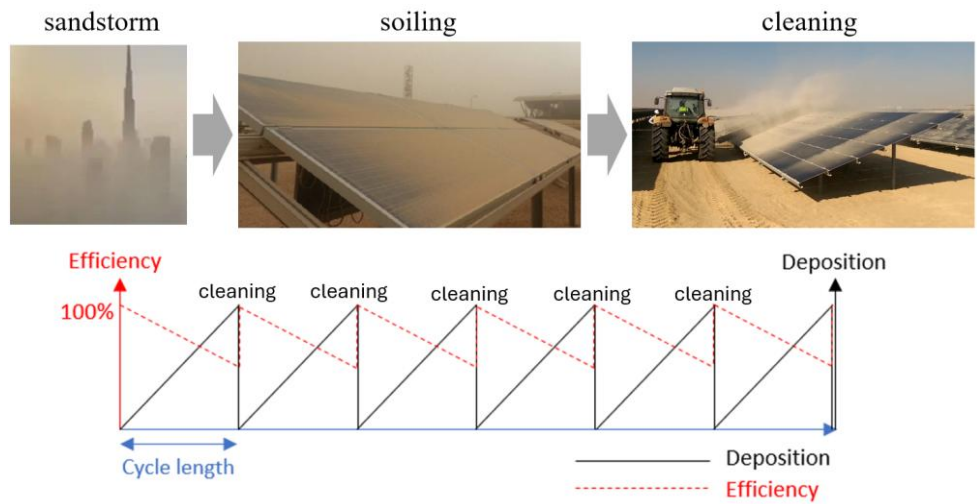

Figure 1. Soiling effect on PV panels

The bottom graph depicts how deposition (solid black line) and efficiency (dashed red line) vary over time. As deposition increases, efficiency drops until a cleaning event occurs, after which efficiency returns to 100%. This cycle repeats over a specified cycle length, emphasizing the impact of soiling and the importance of regular cleaning to maintain optimal performance.

Several studies have highlighted the impact of dust deposition on solar panel efficiency across different regions. For example, a 40% decrease in PV performance was reported after 10 months of the dry season in Saudi Arabia [2], while a 22% loss was observed after one month in India [3]. In Qatar, a 43% reduction was recorded over six months [4]. In the UAE, a 40% efficiency drop was found after three months in Abu Dhabi [5], whereas a 13% decrease was reported over the same period in Sharjah [6].

As a mitigation strategy for dust deposition, cleaning PV panels has been widely adopted using various technologies. However, due to the associated cleaning costs, several studies have conducted economic analyses to determine optimal cleaning frequencies. A cost comparison study for PV cleaning systems in remote sites in the UAE examined the cost components of three cleaning methods [7]. Another study highlighted that the total cost of production losses caused by dust was lower than the total cost of cleaning [8].

A general expression for optimal cleaning intervals was derived and applied to PV modules in Central Saudi Arabia over a year, demonstrating that optimal cleaning costs are smaller than typical operation and maintenance costs [9]. Another study experimentally investigated PV module cleaning scheduling in Kuwait, observing significant efficiency degradation in April, May, and October, where weekly cleaning was necessary to maintain efficiency losses below 15%. For other months, a 30-day cleaning interval was found to be appropriate [10].

A simple formula for optimal PV cleaning periods was developed based on the assumption that dust accumulation increases linearly with time [11]. An improved framework using Sim-Opt was proposed to estimate the economic impact of soiling on PV systems, considering factors such as humidity, tilt angle, and precipitation, and to determine the optimal cleaning frequency [12]. This framework was applied in seven cities globally, including Doha, Qatar, where efficiency losses reached 80% over 140 days. However, limitations included the use of a simple Monte Carlo simulation method with trivial triangular distributions for uncertain factors.

The soiling cost of PV power plants in China was evaluated to range from 16,100 to 22,200 USD/MW, which could be reduced by 36.5-50.3% with the application of a proposed optimal cleaning interval [13]. Based on a model developed after experimental investigation, cleaning frequencies of 10 days and 23 days were recommended for dry and wet cleaning methods, respectively [13]. Similarly, in Al Ain, UAE, the decrease in irradiance intensity and power output for four selected cleaning frequencies was measured. The results showed that output power dropped by 13% for a three-month cleaning interval and only by 4% for a 15-day cleaning interval [6].

A tractor-based cleaning system for PV plants in Senegal was investigated, with an optimal cleaning interval of 14 days determined for the dry season based on two years of data [14]. A deterministic closed-form equation for scheduling was presented, employing the NREL methodology in [15] to account for weather effects. However, the study assumed a constant rate of dust accumulation without incorporating detailed mechanisms and excluded relative humidity in the soiling model, addressing its impact only in post-analysis.

A recent study in [16] explored the economic trade-offs between cleaning costs and energy losses in sustaining solar PV efficiency under weather uncertainty in arid regions. By developing a Sim-Opt framework with a detailed dust accumulation model, they determined an optimal cleaning interval of 34 days for Abu Dhabi. The study also proposed a novel concept of a cleaning subsidy, demonstrating its potential to improve the economic viability of PV systems by altering cost structures beyond traditional metrics such as the levelized cost of electricity. Nonetheless, it could be improved by explicitly incorporating humidity effects and exploring more flexible cleaning schedules.

Other recent studies have increasingly focused on data-driven dynamic cleaning approaches. A two-stage framework for PV cleaning optimization was proposed, where Stage 1 established a baseline cleaning interval using seasonal averages, and Stage 2 dynamically adjusted decisions based on short-term weather forecasts [17]. Using data from a PV system in Shanghai, Monte Carlo simulations evaluated the method under random weather scenarios. However, the study used limited weather variables (radiation, temperature, and cell temperature), a deterministic dust accumulation model fitted seasonally, and simplified normal distribution-based forecasting. Addressing flexibility, an integer linear programming model was developed for cleaning selected portions of PV arrays, incorporating confidence intervals to address uncertainty [18]. Dust deposition was estimated using a regression model in [19] that depended solely on tilt angle, excluding time and weather effects. However, the method lacked the ability to model time-dependent dust accumulation effectively and was limited in its approach to uncertainty, relying solely on confidence intervals.

Further advancing dynamic optimization, two cleaning models were introduced: a deterministic mixed-integer linear programming model and a stochastic Markov Decision Process (MDP) model [20]. The deterministic optimization model was deemed less practical due to weather uncertainties, while the MDP model dynamically updated cleaning decisions using a basic dynamic programming approach. However, limitations included reliance on linear dust accumulation modeling, omission of factors such as wind speed and particulate matter, and the use of a conventional solution method for a simplified MDP model.

A cleaning decision framework was developed based on forecasting the performance ratio of PV systems, where cleaning was triggered when the predicted economic loss from performance ratio degradation equaled the cleaning cost [21]. Initially, multiple variables such as performance ratio, irradiance, active energy, wind speed, temperature, and rainfall were considered, but only performance ratio was included in the final model as the other variables were deemed unhelpful. The forecasting model combined long-term trend prediction using the Prophet library, seasonal trends modeled via Fourier series, and autocorrelation to capture stationary behavior. However, the approach lacked a detailed dust accumulation mechanism and did not incorporate additional data, such as humidity, to improve accuracy.

A dynamic cleaning decision framework was developed, combining deep learning and weather data [22]. Power generation was predicted using an attention-based deep neural network (DNN) trained on weather data, while dust accumulation was estimated using a DNN trained on PV panel images. These predictions, along with daily solar tariff rates and dust accumulation conditions, were input into a support vector machine-coupled DNN to classify whether cleaning was necessary. However, the solar image and weather data were sourced from separate datasets with inconsistencies in location and time alignment. Additionally, weather data were only utilized for power generation predictions, not for cleaning decision-making.

A rolling-horizon framework for PV cleaning recommendations was refined using a two-stage approach [23]. In the first stage, solar radiation and performance were predicted for 90 days using an ensemble long-term and nonlinear autoregressive model. The second stage involved profit estimation and daily cleaning cycle recommendations based on a rolling-horizon strategy. Extensive weather variables, including temperature, cloudiness, fog, haze, hail, rain, sunshine, sandstorm, and snow, were incorporated. However, dust accumulation was modeled solely as a linear function of cover glass type and time, without accounting for weather impacts. The historical dataset also lacked wind speed and humidity, reducing model accuracy.

While prior studies have significantly advanced the understanding of dust accumulation and its impact on solar PV efficiency, notable gaps remain. Most studies model dust accumulation as a linear time-dependent function, with limited exceptions [16], indicating the need for further refinement of dust accumulation models to better capture real-world dynamics. Additionally, essential weather factors such as humidity and wind speed have been either overlooked or only partially considered, despite their potential influence on soiling dynamics and cleaning schedules. Furthermore, while flexible scheduling approaches have been explored, existing methods have significant room for improvement and are often too complex for practical implementation, especially when considering ease of use. These gaps highlight the necessity for advanced modeling techniques and innovative technologies to address these challenges, forming the motivation for this study to develop a more comprehensive and practical solution.

Building on these identified research gaps, this study has two primary objectives. First, it aims to develop an advanced model for dynamic cleaning scheduling using RL, a cutting-edge methodology recognized for its effectiveness in sequential decision-making under uncertainty and its practicality for real-world implementation. Second, it seeks to refine dust accumulation modeling by incorporating additional weather variables, such as humidity and wind speed, while advancing the mechanisms established in the literature. It is important to note that this study specifically targets arid (desert) regions, such as Abu Dhabi, where rainfall is minimal and, therefore, rainfall effects are not considered within the scope of this research. By addressing these challenges, this study contributes to enhancing autonomous energy production system with advanced green technology, promoting more sustainable and efficient solar PV operations in harsh environments. In addition, although the Abu Dhabi government in the UAE recommends a fixed 4-week (28 days) cleaning interval based on empirical evidence [24], evaluating the economic advantages of dynamic scheduling through RL compared to the fixed-interval approach could provide valuable insights.

The rest of this paper is organized as follows. Section 2 outlines the methods used in this study, including environment modeling, Sim-Opt, and RL algorithms. Section 3 presents the results, highlighting key findings and additional analysis for RL decisions. Finally, Section 4 concludes the paper by summarizing the main contributions and suggesting directions for future research.

## 2. Method

The initial phase of this study focuses on gathering relevant data, which is then used to construct a simulation environment that accurately reflects the system's dynamics. Once the environment is modeled, a simulation model is developed and executed to analyze various scenarios. Optimization techniques are subsequently applied to identify the optimal fixed cleaning intervals within given constraints. This Sim-Opt model represents a fixed cleaning scheduling approach under uncertainty and serves as a baseline for performance comparison.

Simultaneously, the dynamic cleaning scheduling problem is formulated as a MDP, with RL methods employed to develop policies that enhance decision-making through iterative interactions with the environment. Finally, the outcomes from both the Sim-Opt and RL approaches are applied to a real-world case study, validating the proposed framework and showcasing its practical applicability and benefits.

The following subsections explain the key components of the research methodology, specifically addressing data collection, environment modeling, Sim-Opt, and RL. The case study is presented in Section 3.

### 2.1. Data Collection

This study collected five key weather parameters from 2018 to 2020 for the Abu Dhabi region, which represents a typical arid climate in the Middle East. Daily data of temperature, wind speed, particulate matter, and irradiance were sourced from the dataset presented in [16], while relative humidity was additionally obtained from visualcrossing.com [25].

**2.2. Environment Modeling**

In environmental modeling, it is crucial to formulate a mathematical relationship that describes PV panel efficiency in relation to soiling, as well as to establish distributions for weather data.

**2.2.1. Soiling modeling**

The daily soiling (*Dsoiling*) in Equation (1) is calculated using the dust accumulation model proposed in [16, 26].

$$Dsoiling = 0.00144 \times (10.6 - 4.99 WS + 247 PM - 73.4 WS \times PM) \quad (1)$$

where *WS* is wind speed in m/s, *PM* is particulate matter in g/m$^2$. Dust may not be entirely cleared by high-speed winds in the UAE due to elevated humidity levels. Therefore, we propose $Dsoiling'$ in Equation (2) which is a calibrated daily soiling considering the effect of relative humidity. The calibration factor, *f(RH)*, is a function of relative humidity *RH*, to represent the portion of dust naturally removed by the wind. If *RH* increases, *f(RH)* decreases due to stickiness.

$$Dsoiling' = \begin{cases} Dsoiling, & Dsoiling \geq 0 \\ f(RH) \times Dsoiling, & Dsoiling < 0 \end{cases} \quad (2)$$

where $f(RH) = k/RH$ and $k = 0.06$. Here, the coefficient *k* is derived from historical data in the UAE and may vary depending on the specific region under consideration.

The soiling is estimated by adding the accumulation up to the previous day and today's soiling as in Equation (3). Assuming that some dust residues would always remain unless being cleaned, the modified soiling at time t, *Soiling(t)'*, is presented in Equation (4) with the minimum residue level $\beta$. $\beta$ is assumed to be 0.01 g/m$^2$ [16].

$$Soiling(t) = Soiling(t-1) + Dsoiling' \quad (3)$$

$$Soiling(t)' = \begin{cases} Soiling, & uncleaned, Soiling \geq \beta \\ \beta, & uncleaned, 0 < Soiling < \beta \\ 0, & cleaned \end{cases} \quad (4)$$

The panel efficiency equation proposed in [27] quantifies the reduction in PV panel efficiency due to soiling. Furthermore, it has been observed that PV panel efficiency gradually decreases over time. To account for this time-dependent degradation, a degradation factor $\tau$ is introduced as shown in Equation (5) and incorporated into Equation (6), which is a modified version of the equation proposed in [27]. For this analysis, we assume an annual degradation rate of 5% [28] and a maximum efficiency of 19.2% for polycrystalline silicon PV panels [29], which is considered for the case study.

$$\tau = (1 - annual\ degradation\ rate)^{\Delta year}, \quad (5)$$

where $\Delta year = current\ year - starting\ year$.

$$Eff(t) = \tau \times (-0.0026 * Soiling(t)'^3 + 0.032 * Soiling(t)'^2 - 0.1369 * Soiling(t)' + 0.192) \quad (6)$$

**2.2.2 Weather data modeling**

In simulation environment, generating random data requires the distribution pattern of the collected data. To identify these patterns, the statistical software Stat::Fit [30] was employed to determine the distribution of each parameter on a month-by-month basis, capturing the unique patterns for each month. Table 1 summarizes these distributions in the format used by Stat::Fit. It is important to note that we selected slightly different distributions compared to those in [16] after careful examinations and the distributions of relative humidity have been presented additionally.

Table 1. Monthly distributions of the collected three-year weather data

| Month | Temperature | Wind Speed | Particulate Matter | Irradiance | Relative Humidity |
|---|---|---|---|---|---|
| 1 | Lognormal (17.0, 1.16, 0.559) | Lognormal (3.77, 1.82, 0.672) | Lognormal (1.56 × $10^{-2}$, −2.78, 1.08) | Weibull(1.93e+03, 4.79, 2.7e+03) | Loglogistic(0, 12, 61.7) |
| 2 | Lognormal (16.0, 1.57, 0.546) | Lognormal (4.05, 1.82, 0.657) | Lognormal (−1.05 × $10^{-2}$, −2.0, 0.796) | Beta(1.61e+03, 6.7e+03, 4.96, 2.23) | Beta(0, 98.8, 10.2, 6.35) |
| 3 | Triangular (18.0, 31.6, 22.2) | Lognormal (2.92, 2.24, 0.467) | Lognormal (−2.04 × $10^{-2}$, −1.86, 0.684) | Johnson SB(-8.48e+03, 1.61e+04, -2.87, 1.13) | Weibull(0, 5.98, 60.3) |
| 4 | Lognormal (13.2, 2.74, 0.178) | Lognormal (2.44, 2.2, 0.351) | Lognormal (−5.17 × $10^{-2}$, −1.56, 0.568) | Beta(1.76e+03, 8.7e+03, 4.39, 1.95) | Beta(0, 77.1, 4.51, 3.15) |
| 5 | Lognormal (−14.1, 3.84, 3.86 × $10^{-2}$) | Lognormal (6.49, 1.47, 0.555) | Lognormal (−1.28 × $10^{-3}$, −1.78, 0.648) | Weibull(5.25e+03, 3.92, 2.39e+03) | Loglogistic(0, 10.3, 45.2) |
| 6 | Normal (34.9, 1.64) | Lognormal (7.21, 1.25, 0.708) | Lognormal (9.11 × $10^{-3}$, −1.64, 0.629) | Weibull(6.5e+03, 4.01, 1.36e+03) | Johnson SB(0, 89, -0.606, 1.89) |
| 7 | Normal (36.1, 2.11) | Lognormal (4.9, 1.87, 0.309) | Lognormal (7.43 × $10^{-2}$, −1.94, 0.811) | Weibull(4.56e+03, 6.48, 2.85e+03) | Johnson SB(0, 82.8, -0.893, 1.98) |
| 8 | Lognormal (29.2, 1.92, 0.125) | Lognormal (−461, 6.16, 3.51 × $10^{-3}$) | Lognormal (2.37 × $10^{-2}$, −1.89, 0.946) | Weibull(4.89e+03, 5.33, 2.22e+03) | Beta(0, 73, 5.98, 1.74) |
| 9 | Normal (34.2, 1.72) | Lognormal (2.56, 2.12, 0.165) | Lognormal (7.96 × $10^{-3}$, −1.97, 0.735) | Weibull(5.4e+03, 4.34, 1.38e+03) | Johnson SB(0, 80.4, -1.19, 1.48) |
| 10 | Normal (30.8, 1.59) | Lognormal (6.25, 1.37, 0.374) | Lognormal (2.9 × $10^{-2}$, −2.76, 0.718) | Weibull(3.38e+03, 7.5, 2.52e+03) | Johnson SB(0, 80.7, -1.43, 1.67) |
| 11 | Triangular (21.3, 29.8, 27.9) | Lognormal (4.88, 1.58, 0.5) | Lognormal (−3.75 × $10^{-3}$, −2.39, 0.694) | Weibull(1.66e+03, 6.85, 3.2e+03) | Weibull(0, 8.48, 61.7) |
| 12 | Normal (22.8, 1.6) | Lognormal (4.89, 1.31, 0.695) | Lognormal (1.11 × $10^{-2}$, −3.03, 0.581) | Weibull(2.38e+03, 7.49, 2.02e+03) | Gamma(0, 59.1, 1.06) |

- Beta(min, max, α₁, α₂): min: min, max: max, α₁: shape 1, α₂: shape 2;
- Gamma(γ, β, α): γ: location, β: scale, α: shape;
- Johnson SB(γ, λ, δ, ξ): γ: location, λ: scale, δ: shape 1, ξ: shape 2;
- Loglogistic(γ, α, β): γ: location, α: scale, β: shape;
- Lognormal(γ, μ, σ): γ: location, μ: scale, σ: shape;
- Normal(μ, σ): μ: mean, σ: standard deviation;
- Triangular(min, max, mode): min: min, max: max, mode: most likely value;
- Weibull(γ, β, α): γ: location, β: shape, α: scale

## 2.3. Simulation and optimization

Figure 2 depicts the overall procedure of the Sim-Opt framework employed in this study. The optimization module seeks to determine the optimal cleaning interval, z, by evaluating values from one up to a specified limit. For each cleaning interval, the optimization module invokes the simulation module, which runs 30 simulation episodes and calculates the average total cost over the PV system's lifetime (20 years). This total cost consists of the sum of the cleaning cost and the energy loss cost both of which depend on the cleaning interval z.

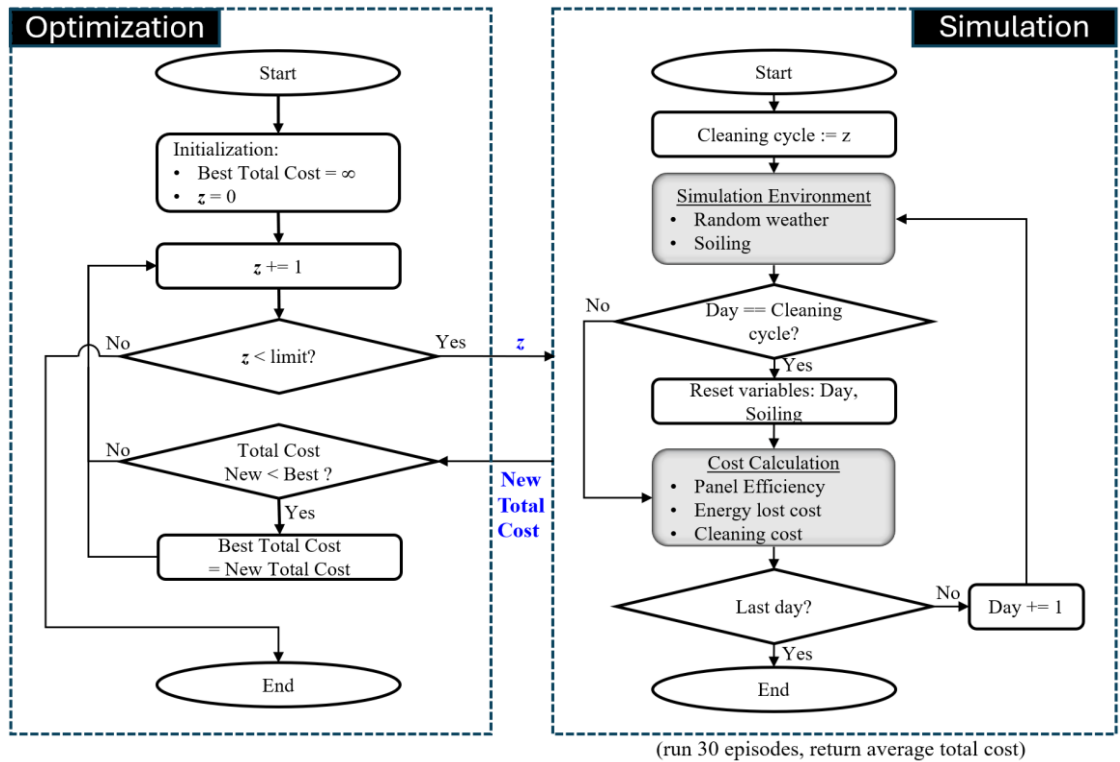

Figure 2. Simulation optimization algorithm

In a single episode of the simulation model, daily random weather conditions are generated, and the resulting soiling is estimated. Based on the cleaning cycle, the soiling level is adjusted, and the corresponding panel efficiency, energy loss cost, and cleaning cost are calculated. This process is repeated incrementally for each day and continues throughout the PV system's lifetime.

**2.4. Reinforcement Learning**

There are many successful RL algorithms, with ongoing research continually enhancing their performance. OpenAI provides a comprehensive taxonomy of RL algorithms, as illustrated in Figure 3 [31]. These algorithms are broadly categorized into model-free and model-based approaches. Within the model-free category, two primary families stand out: policy optimization algorithms and Q-learning algorithms.

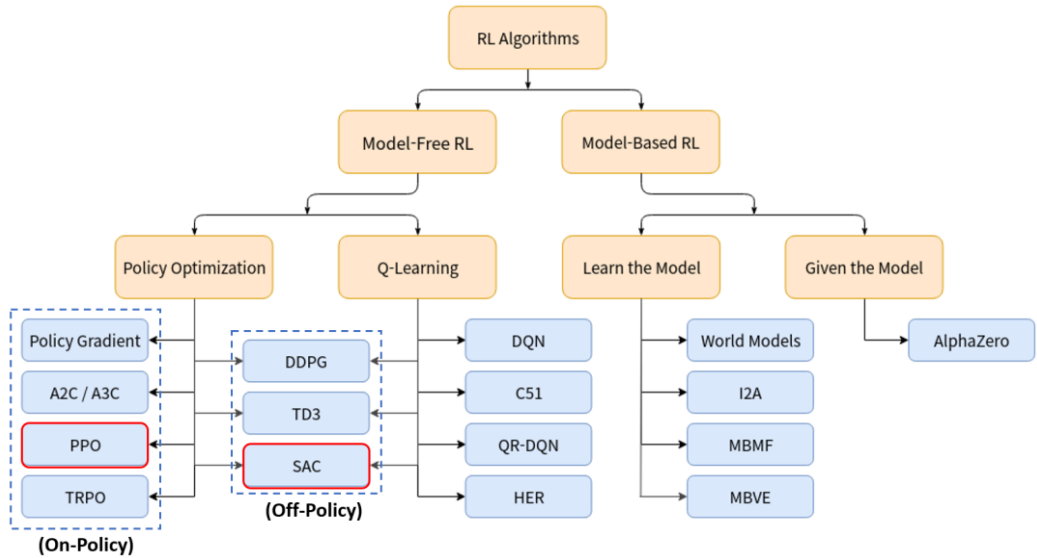

Figure 3. Taxonomy of RL Algorithms by OpenAI [31]

The policy optimization family explicitly represents a policy as $\pi_\theta(a \mid s)$, where a is an action and s is a state. The actor policy selects actions by optimizing its parameters $\theta$ either through direct gradient ascent on the performance objective $J(\pi_\theta)$ or indirectly by maximizing approximations of $J(\pi_\theta)$ in a local context. This optimization process is typically performed on-policy, meaning that each update relies solely on data collected using the current version of the policy. Additionally, policy optimization often incorporates an approximator $V_\varphi(s)$, the critic, which evaluates the policy by estimating the value function $V_\pi(s)$. Among the algorithms in this category, PPO has proven to be one of the most successful.

Q-learning algorithms focus on learning an approximator $Q_\theta(s, a)$ for the optimal action-value function $Q^*(s, a)$, typically using an objective function derived from the Bellman equation. This process is generally performed off-policy, meaning that updates can utilize data collected at any point during training. While Deep Q-Networks (DQN) represent a classic example within this family, SAC builds on DQN by integrating the off-policy training approach with the benefits of policy optimization using stochastic policies. SAC has become one of the most effective algorithms in this category.

Recent research on RL algorithms has been evolving in multiple directions to enhance their applicability in real-world environments, primarily building on strong foundational models such as PPO and SAC. In particular, robust RL has emerged as a key research area to address environmental uncertainties, alongside increasing efforts on adversarial attack defense and safety RL [32]. Multi-agent RL, employing multiple fundamental models, has been widely applied in various domains, including autonomous driving, robotics, and game AI [33]. Additionally, there are ongoing efforts to improve generalization performance through transfer learning and meta-learning [34]. These advancements contribute to transitioning RL from a purely experimental technique to a practical and reliable AI system.

After testing and evaluation of various fundamental candidate RL algorithms, this study selected PPO and SAC for addressing the dynamic cleaning scheduling problem. The following subsections provide a brief overview of the fundamentals of PPO and SAC.

**2.4.1. Proximal Policy Optimization**

PPO, introduced in [35], is an on-policy RL algorithm that builds upon Trust Region Policy Optimization (TRPO) [36]. PPO simplifies the complex Kullback-Leibler divergence constraint in TRPO by introducing a clipping mechanism, leading to a more computationally efficient and stable training process.

It has an actor-critic structure. Its actor maximizes a surrogate objective function using gradient ascent. The objective function is defined as:

$$\max. \hat{\mathbb{E}}_t\left[\min\left(r_t(\theta)\hat{A}_t, clip(r_t(\theta), 1-\epsilon, 1+\epsilon)\hat{A}_t\right)\right] \tag{7}$$

where $r_t(\theta) = \frac{\pi_\theta(a_t|s_t)}{\pi_{\theta_{old}}(a_t|s_t)}$ is the probability ratio, and $\hat{A}_t$ is the truncated generalized advantage estimation (GAE) at timestep $t$:

$$\hat{A}_t = \delta_t + (\gamma\lambda)\delta_t + \cdots + (\gamma\lambda)^{T-1}\delta_{t+T-1} \tag{8}$$

with $\delta_t = r_t + \gamma V(s_{t+1}) - V(s_t)$ and $T$ is the number of learning epochs. Its critic minimizes the mean squared error of the state value function $V(s)$ using gradient descent.

PPO's clipping mechanism constrains policy updates, reducing the risk of training instability and removes the need for complex optimization constraints like Kullback-Leibler divergence, making it computationally lightweight. It is reported that PPO performs well in robotics, gaming, and recommendation systems.

The network structures of the PPO actor-critic are summarized in Table 2. Both networks share the same hidden layer structure, consisting of a fully connected layer (x,256) with ReLU activation. For the actor network, the output layer is a fully connected layer (256,2) with softmax activation, producing a probability distribution over actions (action 0: no-clean, action 1: clean). In contrast, the critic network's output layer is a single scalar value representing the state value estimate $V_\phi(s)$.

Table 2. Actor-critic network structure for PPO and SAC

|  | Component | Actor Network | Critic Network |
|---|---|---|---|
| **PPO** | Input | State (*x*) normalized by scaling (/10) | State (*x*) |
|  | Hidden Layer | Fully connected layer(*x*,256), ReLU activation | Fully connected layer(*x*, 256), ReLU activation |
|  | Output Layer | Fully connected layer(256,2), softmax activation | Fully connected layer(256,1) |
|  | Output | Probability distribution over actions | Scalar value (state value estimate) |
| **SAC** | Input | State (*x*) | Concatenation of state (*x*) and action (*2*) |
|  | Hidden Layer 1 | Fully connected layer (*x*,256), ReLU activation | Fully connected layer (*x+2*,256), ReLU activation |
|  | Hidden Layer 2 | Fully connected layer (256,256), ReLU activation | Fully connected layer (256,256), ReLU activation |
|  | Output Layer | Fully connected layer (256,2), softmax activation | Fully connected layer (256,1) |
|  | Output | Probability distribution over actions | Scalar value (Q-value estimate) |

In this study, the hyperparameters of the PPO algorithm were optimized using Bayesian optimization with Gaussian Processes. The resulting hyperparameters include a learning rate of 0.0005, a discount factor ($\gamma$) of 0.99, a generalized advantage estimation parameter ($\lambda$) of 0.97, a clip range ($\epsilon$) of 0.012, and 5 learning epochs (T).

### 2.4.2. Soft Actor Critic

SAC proposed in [37] is an off-policy RL algorithm that leverages Q-learning principles. SAC introduces entropy regularization to balance exploration and exploitation, maximizing both the expected cumulative reward and the policy's entropy.

SAC also has an actor-critic structure. Its actor maximizes the following objective using gradient ascent:

$$\max. \mathbb{E}_{s_t \sim D, a_t \sim \pi_\theta}\left[Q_\phi(s_t, a_t) - \alpha \log \pi_\theta(a_t|s_t)\right] \quad (9)$$

where $D$ is the replay buffer, and $\alpha$ is a temperature parameter that controls the trade-off between reward maximization and entropy. Its critic minimizes the mean squared error of the Q-function using gradient descent. SAC efficiently utilizes data by maintaining a replay buffer, which enhances sample efficiency and training stability by decorrelating experiences. This feature is particularly beneficial for continuous control tasks, such as robotics control and autonomous driving, where learning effective policies requires diverse and high-dimensional data.

Table 2 also provides a summary of the actor and critic network architectures used in the SAC algorithm. The policy network (actor) takes the state ($x$) as input and processes it through two fully connected layers, each with 256 hidden units and ReLU activation functions. The final output layer, a fully connected layer with softmax activation, produces a probability distribution over possible actions.

The Q-network (critic), on the other hand, takes the concatenation of the state ($x$) and action ($a$) as input. It passes this input through two fully connected layers with 256 hidden units and ReLU activations, followed by an output layer that produces a scalar value representing the Q-value of the given state-action pair. Using this structure, SAC employs a double Q-network to mitigate overestimation bias, using two Q-value estimators to stabilize training and improve the quality of value function updates. This design allows the actor and critic networks to efficiently collaborate in optimizing the policy and estimating value functions.

The hyperparameters for the SAC algorithm were also determined using Bayesian optimization with Gaussian Processes and are as follows: a discount factor ($\gamma$) of 0.99, a target update rate ($\tau$) of 0.005, and an entropy coefficient ($\alpha$) of 0.2, which balances exploration and exploitation. The learning rate for network optimization is set to 0.0003, with a replay buffer size of 100,000 to store past experiences for training. Additionally, a batch size of 256 is used for sampling data from the replay buffer during updates.

## 3. Results

This section presents a design of experiment for a case study in the UAE. Then, the computational results from simulation models and RL models are described. All algorithms in this study were developed and tested on a system equipped with an Intel(R) Core(TM) i9-13900K processor, 128 GB of RAM, and an NVIDIA RTX 4090 GPU, running on Windows 11. The implementation was carried out using Python 3.12 and the PyTorch machine learning library.

Table 3. Test cases of Abu Dhabi for computational experiments [16]

| Electricity Tariff | Case | Tariff (USD/kWh) | Unit Cleaning Cost (USD/Panel/Cycle) |
|---|---|---|---|
| **Expat** | S1exp | 0.073 | 0.0183 |
| | S2exp | 0.073 | 0.0383 |
| | S3exp | 0.073* | 0.0583** |
| | S4exp | 0.073 | 0.0783 |
| | S5exp | 0.073 | 0.0983 |
| **UAE National** | S1uae | 0.018 | 0.0183 |
| | S2uae | 0.018 | 0.0383 |
| | S3uae | 0.018* | 0.0583** |
| | S4uae | 0.018 | 0.0783 |
| | S5uae | 0.018 | 0.0983 |

\* current electricity tariff

\*\*: 0.0583 from (Shah et al. 2020)

### 3.1. Design of experiment

This study adopts the scenarios proposed in [16] to analyze the influence of electricity rates and cleaning costs on cleaning decisions. Table 3 summarizes 10 test cases, featuring two electricity rates: 0.073 USD/kWh for expats in the UAE and 0.018 USD/kWh for UAE nationals. Additionally, five different cleaning costs are considered, including a baseline cost of 0.0583 USD per panel per cycle, along with two higher and two lower values.

**3.2. Simulation results**

For each case listed in Table 3, 30 simulation replications were conducted over a 20-year horizon. The average values of key metrics from these 30 replications were then utilized for subsequent analysis. Figure 4 illustrates the average total cost curves estimated by the simulation model. For expatriate residents (Cases S1–S5exp) in Figure 4 (a), as the unit cleaning cost increases, the total cost curve decreases sharply from a higher initial value, with the minimum total cost rising and shifting to longer optimal cleaning intervals, ranging from 19 to 51 days. Similarly, for UAE nationals (Cases S1–S5uae) in Figure 4 (b), a higher unit cleaning cost also leads to longer optimal cleaning intervals. However, these intervals are significantly longer than those for expats, ranging from 39 to 95 days. Additionally, Figure 4 (b) highlights a flatter total cost curve, suggesting that when electricity tariffs are low, pinpointing the exact optimal cleaning interval becomes less critical.

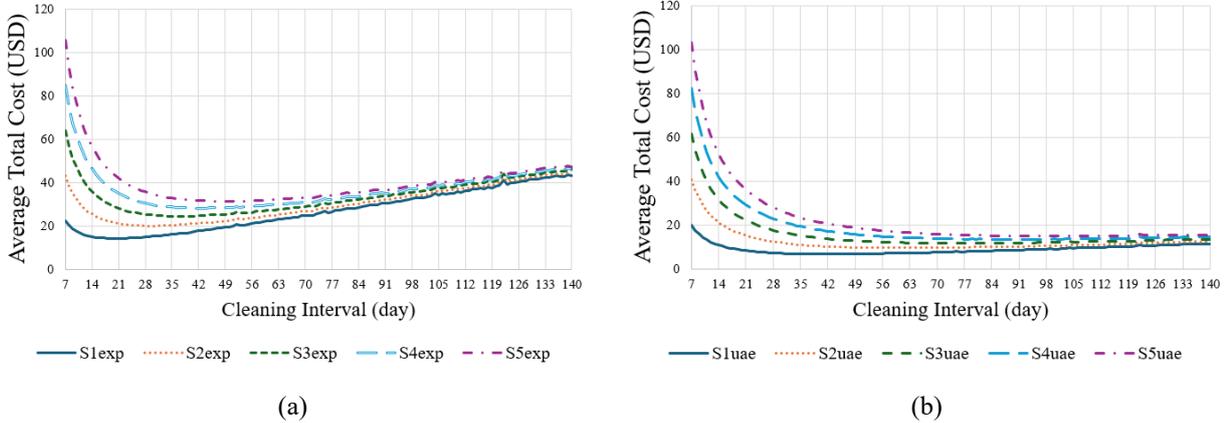

Figure 4. Average total costs by the simulation model

**3.3. Training results of RL models**

To evaluate the performance of PPO and SAC, we trained both algorithms in the same environment under identical conditions. Figure 5 shows the examples of their training behaviors for Case S1exp and highlights distinct differences. Figure 5 (a) illustrates the training progression of PPO. The reward steadily increases during the early episodes and converges to a stable value as training progresses. This behavior demonstrates the inherent stability of PPO, which stems from its policy update mechanism. By restricting the magnitude of policy updates, PPO ensures gradual and consistent improvement, preventing sudden performance degradation. This makes PPO well-suited for environments where stability and reliability are critical.

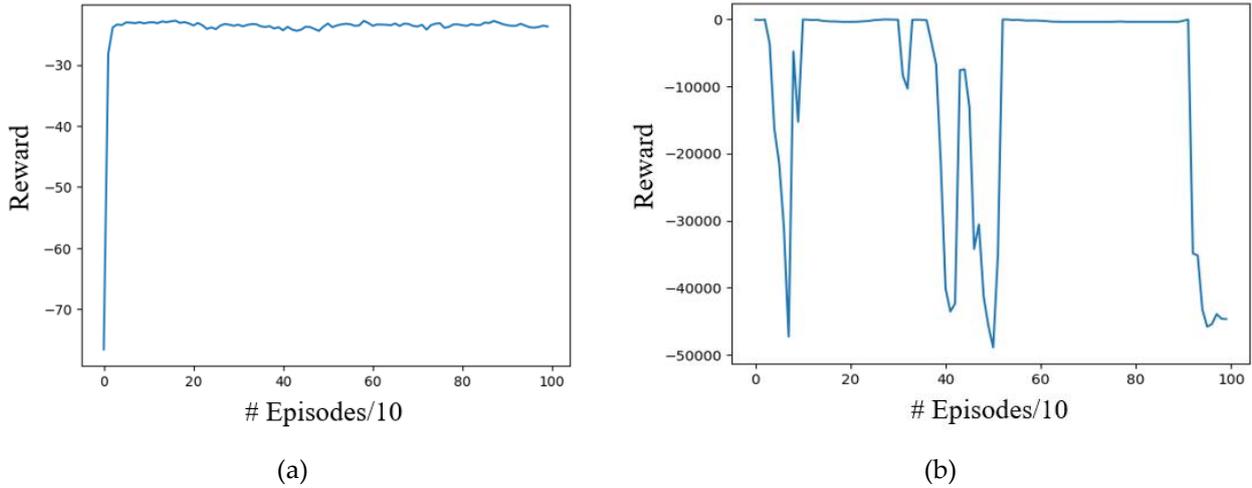

Figure 5. Training results of Case S1exp: (a) PPO and (b) SAC

In contrast, Figure 5(b) shows the training results of SAC, which display significant fluctuations in reward. Although SAC occasionally achieves higher rewards during exploration, it also experiences sharp drops in performance. This instability arises mainly from SAC's off-policy nature and its focus on maximizing policy entropy. While this strategy encourages more exploration and helps find high-reward policies, it also increases the risk of making less effective updates, especially in complex environments with many variables.

Furthermore, when comparing the maximum reward values achieved during training, PPO consistently outperformed SAC, not only in the presented case but also across other test cases. Importantly, SAC failed to outperform Sim-Opt method in all cases. This consistent underperformance underscores the limitations of SAC in this specific application, particularly when compared to the robust and reliable policy optimization capabilities of PPO.

The limited performance of SAC can be explained from several perspectives. First, SAC uses a replay buffer to reuse past data, but in highly random environments, past experiences might not match the current policy well, causing instability in learning [38]. Second, the reward function in our problem has very sparse rewards since we only get rewards after an episode of 20 years. This could make SAC maintain too high entropy, leading to either excessive exploration or insufficient exploration, which can cause instability in the learning process [39]. Third, SAC tries to learn Q-values, but the randomness of the environment might make the estimates unstable, and the sparse reward structure might slow down the updates of the target networks, leading to bigger errors in Q-value estimates [40]. Finally, even though we used Bayesian optimization with Gaussian Processes to tune SAC's hyperparameters, it is possible that the chosen values were not the best, which could negatively affect SAC's performance.

Given these preliminary findings, we elected to use PPO exclusively in the final computational tests. Its superior performance in terms of both reward stability and maximum reward values ensured more reliable and effective policy optimization outcomes.

Overall, the comparison highlights the trade-offs between the two methods. While SAC provides a broader exploration framework, its instability and inability to consistently achieve high rewards make it less suitable in this context. In contrast, PPO offers a robust and stable learning process, making it the more effective choice for achieving and sustaining optimal performance.

### 3.4. Computational results

Table 4 summarizes the computational results for Sim-Opt and PPO. Column 1 lists the case names. Columns 2–4 present the optimal cleaning interval, the average number of cleanings, and the average total cost as determined by Sim-Opt. Columns

5–6 show the average number of cleanings and the average total cost obtained by PPO in testing mode. Column 7 reports the cost savings achieved by PPO compared to Sim-Opt. The final column displays the average total cost recorded by PPO during training mode.

Table 4. Results of Sim-Opt and PPO

| Case | Sim-Opt | | | PPO (testing) | | | PPO (training) |
|---|---|---|---|---|---|---|---|
| | Optimal cycle (days) | Average number of cleanings | (A) Average total cost (USD) | Average number of cleanings | (B) Average total cost (USD) | Cost saving* (%) | Average total cost (USD) |
| S1exp | 19 | 384 | 14.5 | 354 | 12.6 | 13% | 12.5 |
| S2exp | 29 | 251 | 20.4 | 239 | 18.4 | 10% | 18.4 |
| S3exp | 36 | 202 | 24.8 | 221 | 22.9 | 8% | 22.9 |
| S4exp | 41 | 178 | 28.5 | 147 | 26.9 | 6% | 26.4 |
| S5exp | 51 | 143 | 31.7 | 182 | 30.4 | 4% | 29.8 |
| S1uae | 39 | 187 | 6.8 | 189 | 6.3 | 8% | 6.3 |
| S2uae | 60 | 121 | 9.7 | 97 | 9.5 | 2% | 9.2 |
| S3uae | 72 | 101 | 11.7 | 76 | 11.6 | 1% | 11.3 |
| S4uae | 84 | 86 | 13.6 | 64 | 13.8 | -1% | 13.4 |
| S5uae | 95 | 76 | 15.2 | 67 | 14.8 | 2% | 15.1 |

* Cost Saving = (A-B)/A

PPO outperformed Sim-Opt in most cases, achieving cost savings ranging from 1% to 13%, except for Case S4uae. The average number of cleanings prescribed by PPO varied slightly from those determined by Sim-Opt, without a clear pattern emerging. This superior performance of PPO highlights the effectiveness of adapting cleaning schedules dynamically to changing weather conditions, rather than relying on fixed cleaning intervals. This flexibility allows PPO to better respond to the stochastic nature of the environment, optimizing cleaning decisions based on real-time conditions rather than predetermined schedules. While PPO demonstrated excellent performance during training, consistently surpassing Sim-Opt across all cases, its performance in test mode declined in some cases, indicating room for improvement in generalization.

The performance drop observed in PPO's testing mode could be attributed to several factors. First, PPO may have overfitted to the training environment, leading to reduced generalization when exposed to new scenarios in testing. Second, the stochastic nature of the test environment, including variations in weather or other external factors, might differ from the conditions encountered during training, impacting its decision-making. Additionally, the limited exploration during training could result in suboptimal policies when faced with rare or extreme conditions in the test phase. Lastly, the reward structure used in training may not fully capture the complexities of the real-world environment, contributing to discrepancies in performance.

To improve PPO's performance in testing mode, future research could focus on incorporating additional state variables, conducting extensive sensitivity analysis and experimenting with different hyperparameter, and employing more robust training techniques, such as domain randomization [41].

### 3.5. Analysis of PPO decision

To provide a deeper understanding of PPO's dynamic decision-making process, Figure 6 illustrates an example of how cleaning actions are determined. Specifically, it shows the normalized state variables and cleaning decisions for Case S1exp. Figure 6 (a) plots the deposition and days since the last cleaning, variables that seem to have a direct influence on cleaning decisions. These values steadily increase over time until a certain threshold is reached, at which point a cleaning action is triggered.

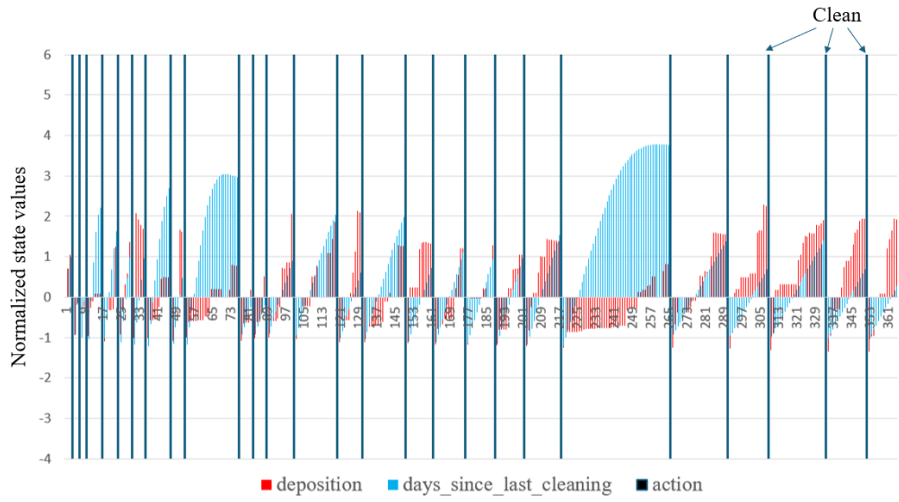

(a)

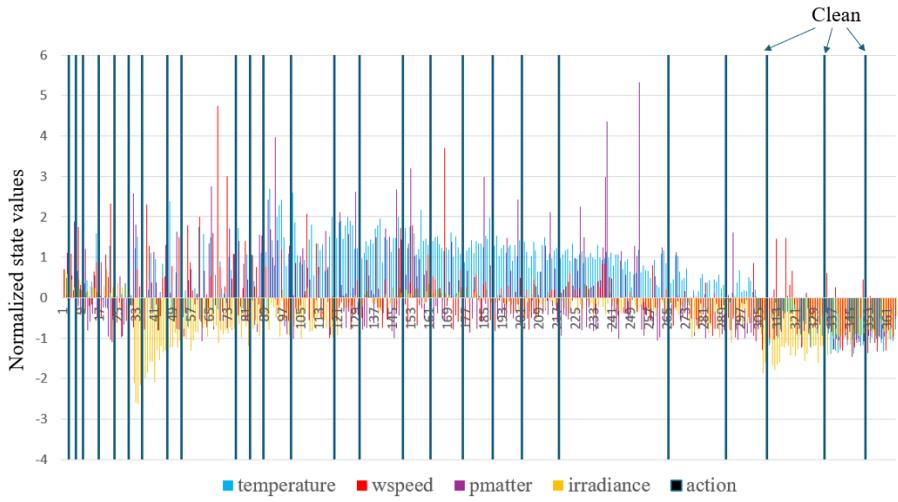

(b)

Figure 6. PPO cleaning decisions with state values (Case S1exp): (a) deposition, days since last cleaning (b) temperature, wind speed, particulate matter, irradiance

On the other hand, Figure 6 (b) displays the remaining state variables: temperature, wind speed, particulate matter, and irradiance. These variables do not exhibit any evident patterns in relation to cleaning decisions, suggesting that they are less likely to directly influence the cleaning actions. This analysis indicates that PPO primarily bases its cleaning decisions on the deposition and days since the last cleaning, while the other variables may have a limited role in this process. Although AI technologies often face challenges in explaining their underlying mechanisms, this analysis, supported by the visualization of state values and cleaning decisions, contributes to the broader field of explainable AI.

## 4. Conclusions

This study developed an RL-based framework to optimize the cleaning schedules of solar PV panels in arid regions. Specifically, the study introduced PPO and SAC algorithms to dynamically adjust cleaning intervals based on environmental conditions. The study also improved existing dust accumulation models by incorporating relative humidity and other weather variables into the soiling mechanism. This approach was validated through a case study in Abu Dhabi, analyzing the economic trade-offs between cleaning costs and energy losses.

The PPO algorithm outperformed both SAC and Sim-Opt approach in most scenarios, achieving cost savings of up to 13% by dynamically adapting cleaning schedules. The results revealed that dynamic scheduling is superior to fixed-interval cleaning, particularly in environments with stochastic weather conditions. The findings also highlighted that deposition and days since the last cleaning were the most influential factors in the RL model's decision-making process.

Despite its effectiveness, PPO demonstrated some limitations, including reduced performance in testing environments, likely due to overfitting and limited exploration of rare environmental conditions during training. The findings underscore the importance of refining RL models to enhance generalization and robustness under varying conditions.

Future research could focus on incorporating additional state variables and advanced hyperparameter tuning techniques to improve the model's generalization capabilities. Exploring alternative RL algorithms and hybrid approaches could provide further insights into balancing exploration and exploitation in dynamic scheduling. Moreover, extending the analysis to include other regions with diverse weather patterns and environmental conditions would enhance the framework's applicability and robustness. Investigating the integration of explainable AI methods into the RL framework would also help stakeholders better understand the decision-making process, facilitating broader adoption of the technology.